%% file: main.tex
\documentclass[preprint,12pt]{elsarticle}

\usepackage{amssymb}
\usepackage{amsmath}
\usepackage{booktabs}

\usepackage{pifont}

\journal{Nuclear Physics B}

\begin{document}

\begin{frontmatter}



% \title{LGD-Net: Latent-Guided Dual-Stream Network for HER2 Expression Prediction via Feature Distillation}
\title{LGD-Net: Latent-Guided Dual-Stream Network for HER2 Scoring with Task-Specific Domain Knowledge}


\author[label1]{Peide Zhu\fnref{eclabel}}

\author[label2]{Linbin Lu\fnref{eclabel}}

\author[label1]{Zhiqin Chen}

\author[label2]{Xiong Chen\corref{cor1}}

\fntext[eclabel]{Equal contribution}
\cortext[cor1]{Corresponding Author}

\affiliation[label1]{organization={School of Mathematics and Statistics, Fujian Normal University},%Department and Organization
            % addressline={ No.18 Middle Wulongjiang Avenue}, 
            city={Fuzhou},
            postcode={350117}, 
            country={China}}

\affiliation[label2]{organization={Department of Oncology, Mengchao Hepatobiliary Hospital of Fujian Medical University}, 
            city={Fuzhou},
            postcode={350117}, 
            country={China}}
%% Abstract
\begin{abstract}

It is a critical task to evalaute HER2 expression level accurately for breast cancer evaluation and targeted treatment therapy selection. However, the standard multi-step Immunohistochemistry (IHC) staining is resource-intensive, expensive, and time-consuming, which is also often unavailable in many areas. Consequently, predicting HER2 levels directly from H\&E slides has emerged as a potential alternative solution. It has been shown to be effective to use virtual IHC images from H\&E images for automatic HER2 scoring. However, the pixel-level virtual staining methods are computationally expensive and prone to reconstruction artifacts that can propagate diagnostic errors. 
To address these limitations, we propose the \textbf{Latent-Guided Dual-Stream Network (LGD-Net)}, a novel framework that employes cross-modal feature hallucination instead of explicit pixel-level image generation. LGD-Net learns to map morphological H\&E features directly to the molecular latent space, guided by a teacher IHC encoder during training. To ensure the hallucinated features capture clinically relevant phenotypes, we explicitly regularize the model training with task-specific domain knowledge, specifically nuclei distribution and membrane staining intensity, via lightweight auxiliary regularization tasks. Extensive experiments on the public BCI dataset demonstrate that LGD-Net achieves state-of-the-art performance, significantly outperforming baseline methods while enabling efficient inference using single-modality H\&E inputs. 

\end{abstract}


% \begin{highlights}
% \end{highlights}

%% Keywords

\end{frontmatter}



% \section{Introduction}
% \label{sec:introduction}
\input{sections/introduction}

% \section{Methodology}
% \label{sec:method}
\input{sections/methodology}

% \section{Experiments}
% \label{sec:experiment}
\input{sections/experiment}

\input{sections/conclusion}
% \section{Results}
% \label{sec:results}
% \input{sections/results}

% \section{Discussion}
% \label{sec:discussion}
% \input{discussion}

% \appendix
% \section{Example Appendix Section}


\bibliographystyle{elsarticle-num} 
\bibliography{reference}

\end{document}

%% file: sections/introduction.tex
\section{Introduction}

Breast cancer is one of the most frequently diagnosed malignancy, imposing a significant burden on public health systems \cite{sung2021global}.  HER2 (Human Epidermal Growth Factor Receptor 2) has been widely adopted as a fundamental biomarker for guiding therapeutic strategies. The accurate assessment of HER2 expression level (categorized as 0, 1+, 2+, or 3+) is  required to identify patients eligible for targeted therapies\cite{waks2019breast,wolff2018human}. % such as Trastuzumab 
The standard clinical workflow typically consists of an initial morphological examination using H\&E (Hematoxylin and Eosin) staining, followed by specific molecular testing via IHC (Immunohistochemistry) \cite{ahn2020her2}. %or Fluorescence In Situ Hybridization (FISH)
This multi-step process is resource-intensive, expensive, and time-consuming.  Furthermore, in many developing regions, IHC infrastructure is often inaccessible. In addition, the subjective interpretation of staining intensity can lead to substantial inter-rator variability \cite{karakas2023interobserver}. Consequently, developing tools that are capable of HER2 scoring automatically directly from H\&E slides has become a critical research objective. 

The rapid advancement of deep learning-based methods has enabled the extraction of complex patterns from histopathological images. Early studies on HER2 scoring directly from H\&E images by training deep neural networks like Convolutional Neural Networks (CNNs) to identify morphological correlates of protein expression \cite{farahmand2022deep,rasmussen2022using}. Although these unimodal approaches established a baseline, they often struggle to distinguish between intermediate HER2 scores (e.g., distinguishing 1+ from 2+), as H\&E staining primarily visualizes tissue structure rather than specific molecular signals \cite{brieu2025auxiliary}. To bridge this information gap, the field has increasingly turned toward \emph{virtual staining} techniques. These methods typically employ generative models such as Generative Adversarial Networks (GANs) to translate H\&E images into synthetic IHC images, effectively creating a proxy modality for downstream classification \cite{bai2023deep}. For instance, Liu et al. \cite{liu2022bci} introduced the BCI dataset and proposed a pyramid-based Pix2Pix model to generate registered IHC pairs. More recently, Qin et al. \cite{qin2025her2} developed a dynamic reconstruction framework that synthesizes missing modalities to improve classification robustness. 

Although these generative approaches have demonstrated the superier performance over unimodal approaches, they rely on pixel-level virtual staining and thus are posed with two major challenges. First, the objective of image translation often diverges from the objective of diagnosis. The generation models trained to minimize pixel-wise reconstruction error or adversarial loss have to learn how to reconstruct details like the high-frequency patterns that are irrelevant to HER2 scoring, such as stromal texture, red blood cells, and background noise. This results in  computational inefficiency and instability during both training and inference. Second, the virtual staining models are prone to generating artifacts or hallucinations in the image space when faced with domain shifts \cite{cohen2018distribution,vasiljevic2022cyclegan}. If a model generates a false positive membrane staining pattern to satisfy the discriminator, the downstream classifier can produce an incorrect diagnosis due to it.

To address these challenges, we introduce a novel method for HER2 expression scoring that avoids explicit image generation in this work. We hypothesize that it is not the \textit{image} of the IHC stain that is required for classification, but rather the \textit{molecular semantics} encoded within it. Therefore, we propose the \textbf{Latent-Guided Dual-Stream Network (LGD-Net)}, a framework that replaces pixel-level translation with \textit{feature hallucination}. Instead of reconstructing a visually interpretable IHC image, LGD-Net learns a mapping from morphological H\&E features to a molecular latent space, guided by a teacher network tht encodes the IHC image to the latent space during training. This approach elliminates the need for the decoder for virtual IHC image generation, focusing the model's capacity on discriminative biomarker features rather than textural reconstruction.

% However,  aligning the high-dimensional hallucinated features with the IHC features can be abstract and physically unconstrained.  
To ensure the hallucinated features remain biologically meaningful, we regualize the the latent space feature generation with the domain knowledge. Specifically, we design lightweight auxiliary tasks that force the latent features to be predictive of nuclei distribution and membrane staining intensity. This ensures that the learned representation is not just a statistical approximation of the teacher network, but explicitly encodes the structural and molecular cues essential for HER2 scoring.

The primary contributions of this paper are as follows:
\begin{itemize}
    \item We propose a cross-modal feature hallucination framework that effectively bridges the gap between H\&E morphology and IHC molecular information, eliminating the computational overhead and artifact risks associated with pixel-level image generation. 
    \item We introduce a domain-knowledge regualization mechanism that enforces physical constraints, specifically, the nuclei density and membrane intensity, on the latent features, improving the model's sensitivity to subtle HER2 expression differences.
    \item We conduct extensive experiments on the public BCI dataset, demonstrating that LGD-Net outperforms both unimodal baselines and recent state-of-the-art generative methods \cite{liu2022bci,qin2025her2}, while requiring only standard H\&E images during inference. 
\end{itemize}

%% file: sections/methodology.tex
\begin{figure}
    \centering
    \includegraphics[width=0.95\linewidth]{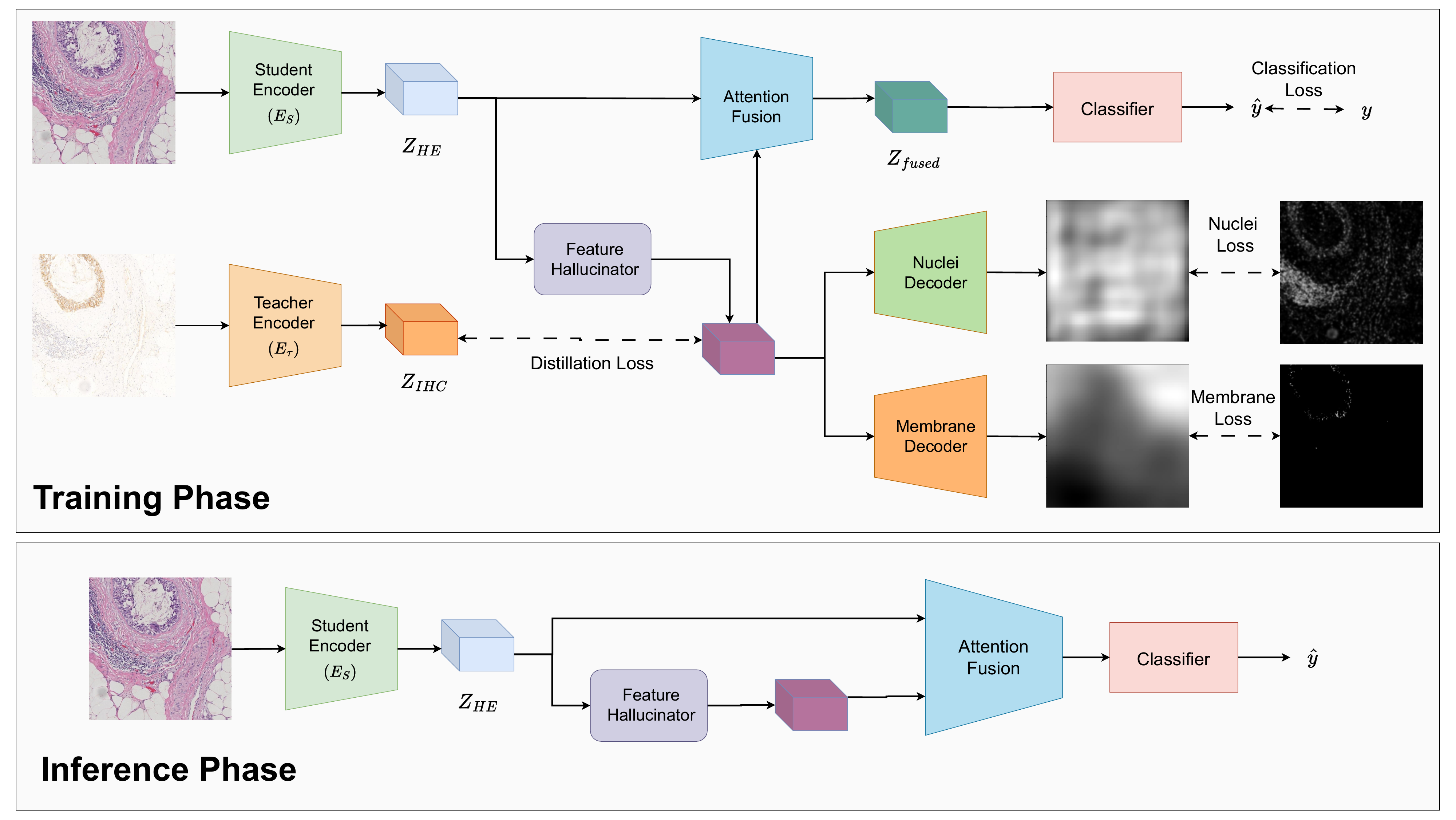}
    \caption{Overview of the proposed LGD-Net framework.}
    \label{fig:architecture}
\end{figure}

\section{Methodology}

We propose the \textbf{Latent-Guided Dual-Stream Network (LGD-Net)}, a framework designed for HER2 expression level scoring from H\&E images. Unlike prevailing virtual staining approaches that depend on computation-intensive  pixel-level translation \cite{liu2022bci,qin2025her2}, LGD-Net employs a \textit{feature hallucination} strategy. This approach enables the inference of missing IHC semantics directly within the latent space without generating the actual visual images, constrained by task-specific pathological priors. % derived from \cite{peng2024advancing}.

\subsection{Problem Formulation and Framework Overview}

Formally, let $\mathcal{D} = \{(x_{HE}^i, x_{IHC}^i, y^i)\}_{i=1}^N$ represent a set of registered H\&E and IHC paired patchs from the BCI cohort \cite{liu2022bci}, where $x_{HE}, x_{IHC} \in \mathbb{R}^{H \times W \times 3}$ are 3-channel RGB images with the height $H$ and width $W$ pixels, and $y \in \{0, 1+, 2+, 3+\}$ denotes the HER2 expression score. Our objective is to learn a predictive model that, \emph{during inference}, accepts only a \emph{single} H\&E input $x_{HE}$ while leveraging the latent molecular information of the absent IHC modality to predict $y$.

As depicted in Fig.~\ref{fig:architecture}, LGD-Net comprises three major components: (1) a \textbf{Dual-Stream Encoding Module} structured in a Teacher-Student configuration; (2) a \textbf{Latent Feature Hallucinator} that maps morphological features to pseudo-molecular representations; and (3) a \textbf{Domain-Knowledge REgularization} module that enforces biological consistency through auxiliary regularization tasks.

\subsection{Cross-Modal Feature Hallucination}

Diverging from conventional image-to-image translation, we perform cross-modal synthesis exclusively within the high-level feature space. In this way, in stead of generating high-frequency artifacts, we focus on the model's capacity on discriminative biomarkers.  

\subsubsection{Dual-Stream Encoders}
We adopt a parallel encoder architecture: a Student Encoder $\mathcal{E}_{S}$ processing H\&E inputs and a Teacher Encoder $\mathcal{E}_{T}$ processing IHC inputs. Leveraging the feature extraction protocol from \cite{qin2025her2}, we obtain the latent representations:
\begin{equation}
    z_{HE} = \mathcal{E}_{S}(x_{HE}), \quad z_{IHC} = \mathcal{E}_{T}(x_{IHC})
\end{equation}
where $z \in \mathbb{R}^{C \times h \times w}$ denotes the spatial feature maps. During training, $\mathcal{E}_{T}$ provides the target distribution for molecular feature learning; critically, it is discarded during the inference phase, requiring only H\&E input.

\subsubsection{Latent Mapping}
To bridge the domain gap between the H\&E and IHC modalities, we introduce the Feature Hallucinator $\mathcal{M}$. This module learns a non-linear transformation to project H\&E features into a ``hallucinated'' IHC latent space:
\begin{equation}
    \hat{z}_{IHC} = \mathcal{M}(z_{HE})
\end{equation}
The primary objective is to align $\hat{z}_{IHC}$ structurally and semantically with the authentic IHC features $z_{IHC}$  without pixel reconstruction.

\subsection{Domain Knowledge-constrained Training via Auxiliary Tasks}

Minimizing statistical discrepancy between latent vectors is insufficient to guarantee that the hallucinated features capture clinically relevant phenotypes.  The characterization of \textit{nuclei distribution} and \textit{membrane staining intensity} is critical for accurate HER2 assessment. Therefore, we propose to internalize these biological constraints into the latent representation via two lightweight auxiliary decoders. 

\subsubsection{Nuclei Density Regularization} 
To ensure $\hat{z}_{IHC}$ retains accurate structural information, we incorporate a Nuclei Decoder $\mathcal{D}_{nuc}$. Ground truth density maps $K_{gt}$ are generated from the Hematoxylin channel of the training images using color deconvolution and Gaussian filtering \cite{peng2024advancing}. The decoder estimates this distribution as:
\begin{equation}
    \hat{K} = \mathcal{D}_{nuc}(\hat{z}_{IHC})
\end{equation}
By minimizing the reconstruction error between $\hat{K}$ and $K_{gt}$, we constrain the latent features to encode precise cellular localization.

\subsubsection{Membrane Intensity Regularization}
Given that HER2 scoring is principally defined by the completeness of membrane staining (DAB signal), we employ a Membrane Decoder $\mathcal{D}_{mem}$ to predict membrane activation. The ground truth mask $M_{gt}$ is derived via adaptive thresholding of the DAB channel in the HED color space \cite{peng2024advancing}. The predicted mask is formulated as:
\begin{equation}
    \hat{M} = \mathcal{D}_{mem}(\hat{z}_{IHC})
\end{equation}
These auxiliary tasks ensure that $\hat{z}_{IHC}$ is not merely an abstract vector but explicitly encodes the biological semantics required for the diagnostic task.

\subsection{Dynamic Fusion and Classification}

For the final prediction, we integrate the morphological features $z_{HE}$ with the hallucinated molecular features $\hat{z}_{IHC}$. To account for potential variations in the reliability of the hallucinated features, we implement the modality-sensitive feature attention-based fusion mechanism.  

We concatenate the feature maps and compute a spatial-channel attention map $\mathcal{A}$ via shared MLP layers:
\begin{equation}
    z_{fused} = \text{Concat}(z_{HE}, \hat{z}_{IHC}) \cdot \mathcal{A}(z_{HE}, \hat{z}_{IHC})
\end{equation}
The fused representation is then fed into a classifier $\mathcal{C}$ to compute the probability distribution over HER2 expression levels:
\begin{equation}
    \hat{y} = \mathcal{C}(z_{fused})
\end{equation}

\subsection{Loss Functions and Optimization}

To train the proposed model, we design a composite function to collaboratively optimize the classification performance while enforcing cross-modal alignment and biological consistency, specifically with the following training objectives:

\begin{enumerate}
    \item \textbf{Classification Loss ($\mathcal{L}_{cls}$):} We utilize a Cross-Entropy loss:
    \begin{equation}
        \mathcal{L}_{cls} = -\sum_{c} y_{c} \log(\hat{y}_{c})
    \end{equation}
    %w_c 

    \item \textbf{Feature Distillation Loss ($\mathcal{L}_{dist}$):} To align the hallucinated and real IHC features, we minimize the cosine distance, prioritizing semantic direction over magnitude: 
    \begin{equation}
        \mathcal{L}_{dist} = 1 - \frac{\hat{z}_{IHC} \cdot z_{IHC}}{\|\hat{z}_{IHC}\|_2 \|z_{IHC}\|_2}
    \end{equation}

    \item \textbf{Auxiliary Biological Losses ($\mathcal{L}_{bio}$):} These losses enforce the domain knowledge constraints. We apply Mean Squared Error (MSE) for nuclei density regression and Dice Loss for membrane segmentation:
    \begin{equation}
        \mathcal{L}_{nuc} = \|\hat{K} - K_{gt}\|_2^2, \quad \mathcal{L}_{mem} = 1 - \text{Dice}(\hat{M}, M_{gt})
    \end{equation}
\end{enumerate}

The total optimization objective is then formulated as:
\begin{equation}
    \mathcal{L}_{total} = \mathcal{L}_{cls} + \lambda_{d} \mathcal{L}_{dist} + \lambda_{n} \mathcal{L}_{nuc} + \lambda_{m} \mathcal{L}_{mem}
\end{equation}
where $\lambda_{d}, \lambda_{n}, \lambda_{m}$ are hyperparameters balancing feature alignment and biological regularization. Upon completion of training, the auxiliary decoders are removed, resulting in a streamlined inference architecture that requires only H\&E input.

%% file: sections/experiment.tex
\begin{figure}[h]
    \centering
    \includegraphics[width=0.55\linewidth]{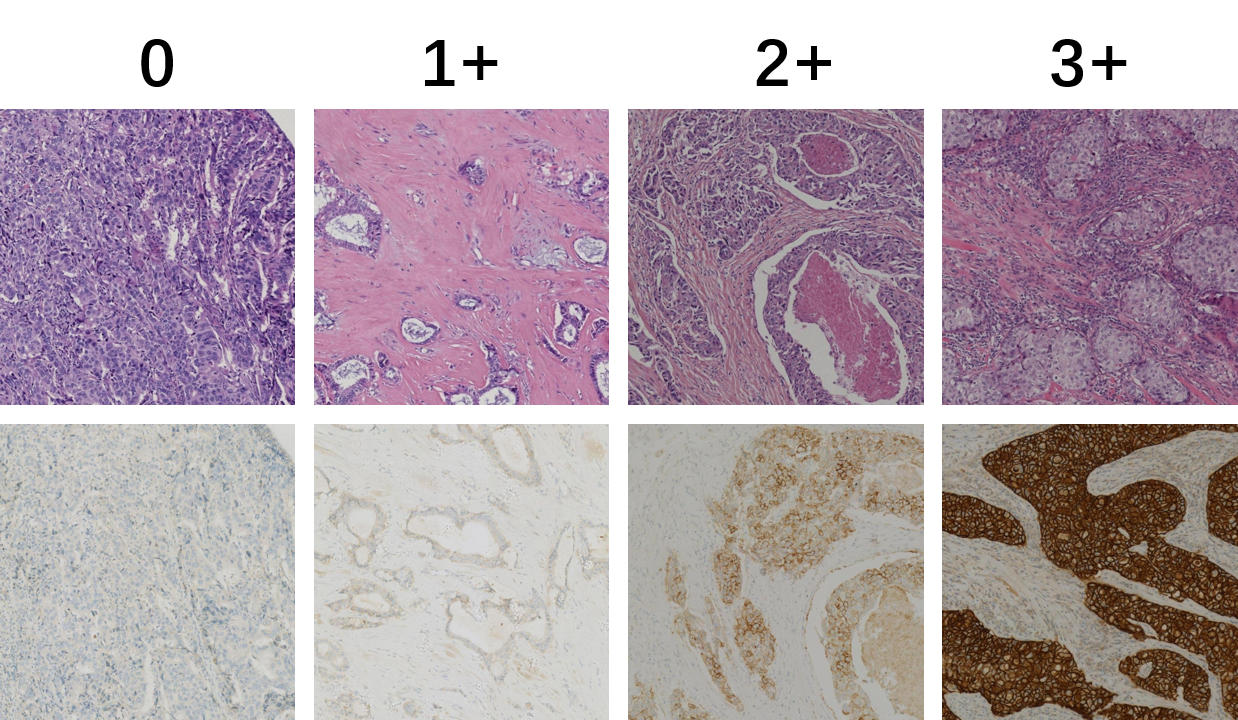}
    \caption{ Examples of paired H\&E-IHC images with different
HER2 expression levels (0, 1+, 2+, 3+ from left
to right). }
    \label{fig:example}
\end{figure}

\section{Experimental Setup}

\subsection{Dataset}

We conducted experiments to validate the LGD-Net on the \textbf{BCI Dataset} \cite{liu2022bci}, a large publicly available cohort for HER2 assessment. The dataset comprises 4,873 pairs of H\&E and IHC patches ($1024 \times 1024$ pixels)  derived from strictly spatially registered whole slide images with the HER2 expression level annotations ranging from negative (0) to strongly positive (3+). 
% What makes the BCI dataset unique and valuable is that each pair of patches has been annotated with its standard HER2 expression level.   
We use the official training and test sets (3896:977 images) for training and testing.

To generate ground truth supervision for the auxiliary tasks, we employed the deterministic image processing pipeline described in \cite{peng2024advancing}. Specifically, nuclei density maps ($K_{gt}$) were generated by applying color deconvolution to extract the hematoxylin channel, followed by Gaussian filtering.  
Membrane staining masks ($M_{gt}$) were derived via adaptive thresholding of the DAB channel in the Haematoxylin-Eosin-DAB (HED) color space.

\subsection{Implementation Details}
We implemented the propsoed framework in PyTorch.  The model was trained on a server with 256GB memory and two NVIDIA RTX 3090 GPU. We utilized ResNet-50 \cite{he2016deep}, initialized with ImageNet weights, as the backbone for both the Student and Teacher encoders. Input images were resized to $512 \times 512$ pixels to optimize the trade-off between computational efficiency and fine-grained feature retention. 
To generate the fake IHC images for baselines, we trained the generator with an H800 GPU. 

We conduct model training was performed in an end-to-end manner using the AdamW optimizer ($\beta_1=0.9, \beta_2=0.999$, weight decay=$1e-4$). We initialized the learning rate at $1e-4$ and applied a cosine annealing schedule over 50 epochs. The loss weighting factors were empirically set to $\lambda_{d}=10.0$, $\lambda_{n}=5.0$, and $\lambda_{m}=5.0$ to prioritize feature alignment and biological regularization. 

\subsection{Evaluation Metrics}
We report the \emph{Accuracy (Acc)} and \emph{Macro-averaged F1-score (F1)} to assess classification performance,  with the latter being particularly critical given the inherent class imbalance. Additionally, we also report the \emph{Cohen's Kappa ($\kappa$)} to quantify the agreement between model predictions and ground-truth annotations.

\section{Results and Analysis}
\subsection{Comparison with State-of-the-Art Methods}

To validate the efficacy of LGD-Net, we compared it against the following categories of baseline approaches that cover most HER2 prediction methods: 
\begin{enumerate}
    \item \textbf{Unimodal Baseline:} A standard ResNet-50 trained exclusively on single modality ( H\&E images or IHC images only). 
    \item \textbf{Dual-modal Baselines:} For this category of baselines, we consider three settings. The first is \emph{image concat}, which takes both the paired H\&E image and IHC image as the input to the convolutional layer.  The second is \emph{feature level concat with attention} which encoder the H\&E image and the IHC image separately, and apply attention on the concated features. The third is the \emph{Feature-Level Fusion}, where we use the Bidirectional Cross-Modal Reconstruction framework \cite{qin2025her2}, which represents the current state-of-the-art. 

\end{enumerate}

\begin{table}[h]
    \centering
    \caption{Quantitative comparison on the BCI testing set..}
    \label{tab:sota}
    \begin{tabular}{l|ccc}
        \hline
        Method  & Acc(\%,$\uparrow$) & F1 ($\uparrow$) & $\kappa$ ( $\uparrow$) \\
        \hline
        H\&E Unimodal  & 82.29 & 0.7935 & 0.7543 \\
        IHC Unimodal   & 89.46 & 0.8906 & 0.8642 \\
        \hline 
        ImageConcat ( H\&E + real IHC ) & 90.99 & 0.8929 & 0.8724 \\
        FeatureConcat ( H\&E + real IHC) & 93.76 & 0.9429 & 0.9208 \\
        FeatureFusion (H\&E + real IHC) & 94.37 & 0.9375 & 0.9190 \\

        \hline
        \textbf{LGD-Net (Ours)} (H\&E) & \textbf{95.60} & \textbf{0.9644} & \textbf{0.9453} \\
        \hline
    \end{tabular}
\end{table}

As reported in Table~\ref{tab:sota}, the H\&E-Only baseline yields suboptimal performance (82.29\%), confirming that morphological features alone are insufficient to resolve ambiguous HER2 classes. The IHC-only baseline outperforms H\&E baseline as expected since the IHC images directly contain the visual features of the HER2 expression.  
The dual-modal methods improve accuracy by synthesizing the H\&E and IHC information. We can also observe that the feature-based methods including the attention on the concatenated features and the advanced feature fusion can further improve the performance. 
% they are computationally intensive and limited by reconstruction artifacts. 
Notably, LGD-Net surpasses the previous state-of-the-art method by 1.23\% in accuracy, 0.0269 in Macro-F1 and 0.0263 in Kappa. This performance gain corroborates our hypothesis that \textit{feature hallucination}, constrained by internalized domain knowledge, captures diagnostic semantics more effectively than pixel-level features.

\subsection{Ablation Study}

We conducted a comprehensive ablation study to investigate the contribution of each component within LGD-Net. The results are detailed in Table~\ref{tab:ablation}.

\begin{table}[h]
    \centering
    \caption{Ablation study of key components: Feature Hallucination (Halluc.), Attention Fusion (Attn.), and Domain Knowledge Internalization (Bio-Reg).}
    \label{tab:ablation}
    \begin{tabular}{l|ccc|ccc}
        \hline
        Variant & Halluc. & Attn. & Bio-Reg & Acc (\%) & F1 & $\kappa$ \\
        \hline
        A (Baseline) & \ding{55} & \ding{55} & \ding{55} & 82.29 & 0.7935 & 0.7543 \\
        B (Feat. Align) & \ding{51} & Concat & \ding{55} & 92.53 & 0.9120 & 0.8914 \\
        C (Dynamic) & \ding{51} & \ding{51} & \ding{55} & 93.35 & 0.9344 & 0.8884   \\
        D (w/ Nuclei) & \ding{51} & \ding{51} & Nuclei & 94.27 & 0.9498  &  0.9239 \\
        E (w/ Memb.) & \ding{51} & \ding{51} & Memb. & 94.78
        & 0.9522  & 0.9281  \\
        \hline
        \textbf{F (Full)} & \ding{51} & \ding{51} & \textbf{Both} & \textbf{95.60} & \textbf{0.9644}  &  \textbf{0.9453}\\
        \hline
    \end{tabular}
\end{table}

\textbf{Efficacy of Feature Hallucination:} Simply introducing the feature hallucination module (Variant B) results in a substantial performance improvement over the baseline (+10.24\% in accuracy, 0.1371 in $\kappa$). This indicates that the student encoder and the feature hallucinator successfully learns to approximate the molecular feature distribution of the Teacher.

\textbf{Impact of Domain Knowledge:} The integration of biological constraints further enhances robustness. Specifically, membrane intensity regularization (Variant E) yields slightly heigher improvement than nuclei regularization (Variant D). This aligns with clinical guidelines, which identify membrane staining completeness as the primary determinant for distinguishing HER2 scores \cite{peng2024advancing}. The combination of both constraints (Variant F) achieves the highest performance, demonstrating that structural (nuclei) and molecular (membrane) cues provide complementary diagnostic information. 

% \subsection{Qualitative Analysis}

%% file: sections/conclusion.tex
\section{Conclusion}

In this paper, we presented the \textbf{Latent-Guided Dual-Stream Network (LGD-Net)}, a novel framework for accurate HER2 scoring directly from H\&E stained images. To address the inherent limitation of H\&E images of the lack of molecular information required to distinguish HER2 representation levels, we proposed to integrate feature-level hallucination instead of pixel-level virtual staining.  
The proposed approach demonstrates that by hallucinating molecular features in the latent space and constraining them with task-specific domain knowledge, including the nuclei density and membrane staining intensity, LGD-Net effectively recovers diagnostic signals lost in standard morphological stains. 

Comprehensive experiments on the BCI dataset validate the effectiveness of our method. LGD-Net achieved a significant performance gain over the unimodal H\&E baseline (improving accuracy from 82.29\% to 95.60\% ) and outperformed state-of-the-art dual-modal fusion methods. 
Furthermore, by eliminating the computationally expensive image decoder during inference, LGD-Net offers a highly efficient solution suitable for large-scale clinical screening. 